\documentclass{article}

 \PassOptionsToPackage{numbers, compress}{natbib}

\usepackage[dblblindworkshop,final]{neurips_2025}

\usepackage[utf8]{inputenc} 
\usepackage[T1]{fontenc}    
\usepackage{url}            
\usepackage{booktabs}       
\usepackage{amsfonts}       
\usepackage{nicefrac}       
\usepackage{microtype}      
\usepackage{xcolor}         
\usepackage{graphicx}
\usepackage{tabularx}
\usepackage{subcaption}
\usepackage[symbol]{footmisc}
\usepackage[inline]{enumitem}
\usepackage{amsmath}
\usepackage{fontawesome5}
\usepackage{bbding}
\definecolor{freezingblue}{RGB}{100, 160, 255}
\definecolor{burningred}{RGB}{255, 90, 0}
\workshoptitle{Efficient Reasoning}

\title{Attention Guided Alignment in Efficient
Vision-Language Models
}

\author{Shweta Mahajan \footnotemark[1]  \quad  Hoang Le \footnotemark[1]  \quad  Hyojin Park \quad  Farzad Farhadzadeh \\[1pt] \textbf{Munawar Hayat} \quad \textbf{Fatih Porikli}\\
Qualcomm AI Research\footnotemark[2]\\
\texttt{\{shwemaha,hoanle,hyojinp,ffarhadz,hayat,fporikli\}@qti.qualcomm.com}
}

\begin{document}

\maketitle

\begin{abstract}
Large Vision-Language Models (VLMs) rely on effective multimodal alignment between pre-trained vision encoders and Large Language Models (LLMs) to  integrate visual and textual information. 
This paper presents a comprehensive analysis of attention patterns in efficient VLMs, revealing that concatenation-based architectures frequently fail to distinguish between semantically matching and non-matching image-text pairs.
This is a key factor for object hallucination in these models.
To address this, we introduce Attention-Guided Efficient Vision-Language Models (AGE-VLM)—a novel framework that enhances visual grounding through interleaved cross-attention layers to instill vision capabilities in pretrained small language models.
This enforces in VLM the ability "look" at the correct image regions by leveraging spatial knowledge distilled from the Segment Anything Model (SAM), significantly reducing hallucination.
We validate our approach across different vision-centric benchmarks where our method is better or comparable to prior work on efficient VLMs.
Our findings provide valuable insights for future research aimed at achieving enhanced visual and linguistic understanding in VLMs.

\end{abstract}

\section{Introduction}
\label{sec:intro}
\footnotetext[1]{Equal Contribution.}
\footnotetext[2]{Qualcomm AI Research is an initiative of Qualcomm Technologies,
Inc.}

Large Vision-Language Models (VLMs) \cite{alayrac2022flamingo,liu2023visual,radford2021learning,tong2024cambrian,wang2024qwen2,zhu2023minigpt} leverage the capabilities of pre-existing Large Language Models (LLMs) \cite{achiam2023gpt,chowdhery2023palm,grattafiori2024llama} to address complex tasks. Although LLMs excel in text-only domains such as natural language understanding \cite{grattafiori2024llama}, mathematics \cite{cobbe2021training}, and coding \cite{le2022coderl} by following task-specific instructions, VLMs extend these abilities to the multimodal realm. This enables them to perform tasks like image understanding, captioning, object localization, and multi-turn visual question answering. Architecturally, VLMs typically consist of three key components: a vision encoder to process visual input, an adapter to map visual representations into the language model's token space, and a decoder-only LLM that processes these combined representations. The fusion of visual and textual information is commonly achieved either by concatenating visual tokens with text tokens for processing by self-attention layers or by interleaving visual tokens using dedicated cross-attention layers within the LLM \cite{ alayrac2022flamingo,grattafiori2024llama}.

\begin{figure}
    \centering
    \includegraphics[width=\linewidth]{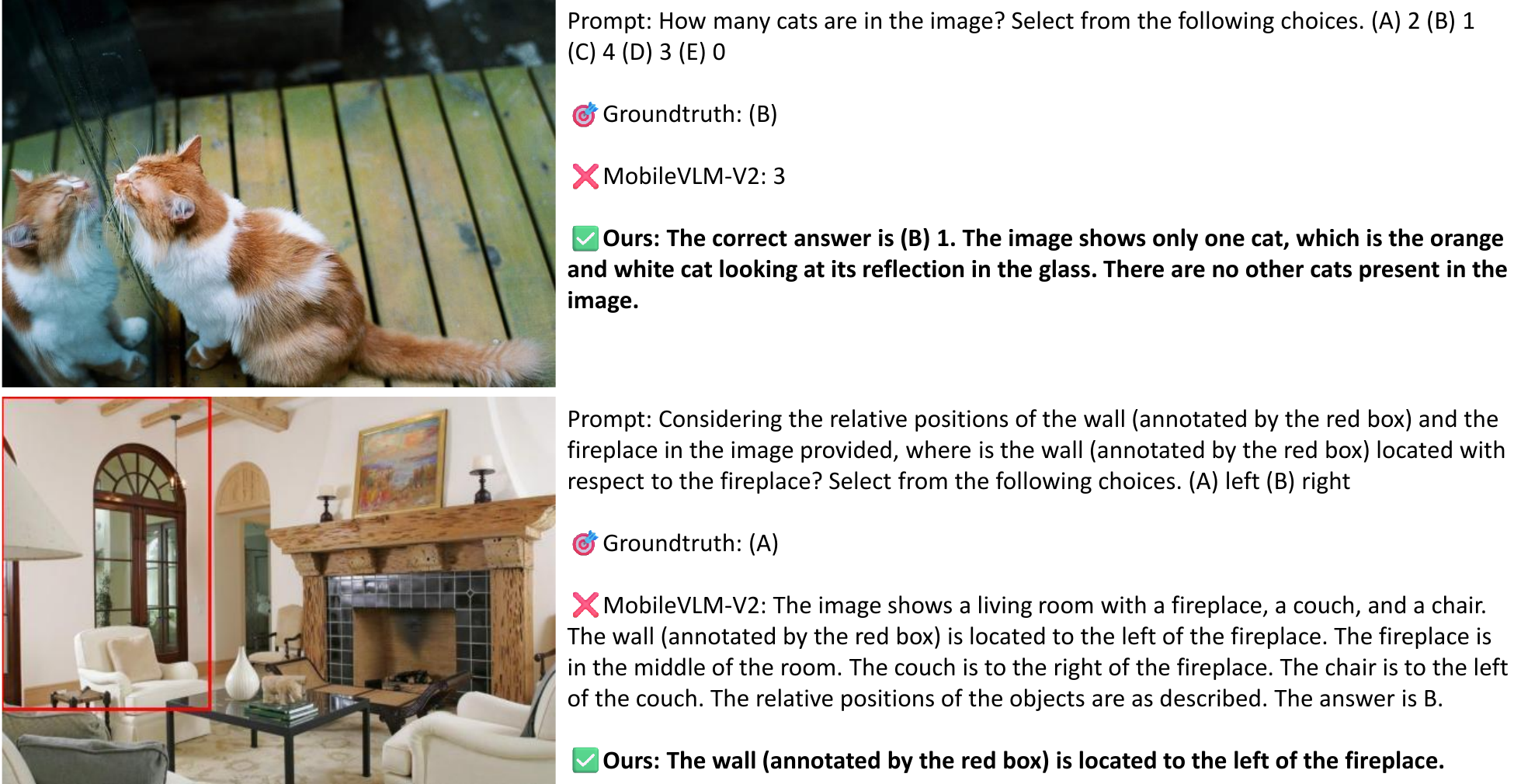}
    \caption{\textbf{Object hallucination and mitgation in efficient vision language models.} We show that prior work on efficient VLMs fails to localize (top) and correctly count the number of objects (cat) in images. Our attention guided efficient VLMs approach equipped with the knowledge distillation from Segment Anything Model in the cross-attention weights can effectively generate correct responses with explanation grounded in the visual domain.}
    \label{fig:teaser}
\end{figure}

Recent research \cite{rahmanzadehgervi2024vision,tong2024cambrian,tong2024eyes} has shown a significant challenge in VLMs: a tendency to ignore visual modality representations when performing vision-language tasks. These models may produce answers, whether correct or incorrect, relying solely on textual instructions and associated questions, thereby ignoring crucial visual information.  \citeauthor{tong2024cambrian}~ \cite{tong2024cambrian} observe that the performance gap with and without visual information is  less than a 5\% on multiple benchmarks including MMMU \cite{yue2024mmmu}, MathVista \cite{lu2023mathvista}, and AI2D \cite{hiippala2021ai2d}. To mitigate this, various efforts have focused on enhancing visual capabilities, including curating vision-centric datasets \cite{tong2024cambrian} and improving vision-text alignment through auxiliary mechanisms such as specialized projection mechanisms \cite{masry2025alignvlm}. However, these limitations often persist even in efficient frameworks designed for smaller-scale models with fewer parameters. Despite such enhancements, many multimodal approaches still struggle with effective visual information processing,  exhibiting issues like object hallucination \cite{guan2024hallusionbench} as shown in Fig.~\ref{fig:teaser}.

Many efficient VLMs \cite{ge2024convllava,vasu2024fastvlm} employ convolutional vision encoders like ConvNeXt \cite{liu2022convnet}. 
To enable vision-language capabilities, ConvNeXt is trained contrastively with CLIP \cite{ge2024convllava}. 
However, this approach leads to lack of fine-grained spatial grounding in the visual features.
During standard VLM training, given the LLM's strong language prior and the use of next-token prediction with cross-entropy loss, the model fails to recover the fine-grained visual information.
While larger models \cite{lin2023sphinx,wang2024qwen2} can integrate features from multiple encoders (e.g., DINO \cite{oquab2023dinov2} , ViT\cite{dosovitskiy2020image}) to improve grounding, this is unfeasible for resource-constrained VLMs.

To endow the efficient models with spatial grounding of the vision features and to mitigate object hallucination, we propose a novel framework called \emph{Attention-Guided Efficient VLM} (AGE-VLM). 
Our approach modifies a standard LLM by interleaving cross-attention layers with its existing self-attention layers. 
The core idea is to distill spatial knowledge from the Segment Anything Model (SAM) \cite{kirillov2023segment} directly into these cross-attention mechanisms. This is achieved by optimizing the cross-attention weights to align with segmentation masks generated by SAM for relevant text queries. Consequently, the VLM learns to "look" at the correct regions of interest when processing multimodal inputs.
A key advantage is the data efficiency of this distillation process, enabling enhanced grounding with limited training examples.
We make following contributions: 
\begin{itemize}
    \item We analyze the vision and text features in the hidden states of the efficient VLMs and uncover their limitations in disambiguating the semantics between similar and dissimilar image-text  pairs to uncover limitations in vision-centric tasks including object hallucination.
    \item To endow relatively small LLMs (1B parameter models) with vision capabilities in an efficient manner, we propose a new efficient multimodal framework with cross-attention layers which  leverage attention-guidance from segmentation model (SAM).
    \item  To distill knowledge from the SAM model, we introduce a four stage training paradigm  which seamlessly integrates the vision features with the pretrained LLM without effecting the language capability of the underlying model. Our efficient AGE-VLM with guidance loss only during pretraining stage outperforms prior art across vision-centric tasks.
\end{itemize}

\begin{figure}[t!]
    \centering
    \begin{subfigure}[t]{0.5\textwidth}
        \centering
        \includegraphics[width =\linewidth]{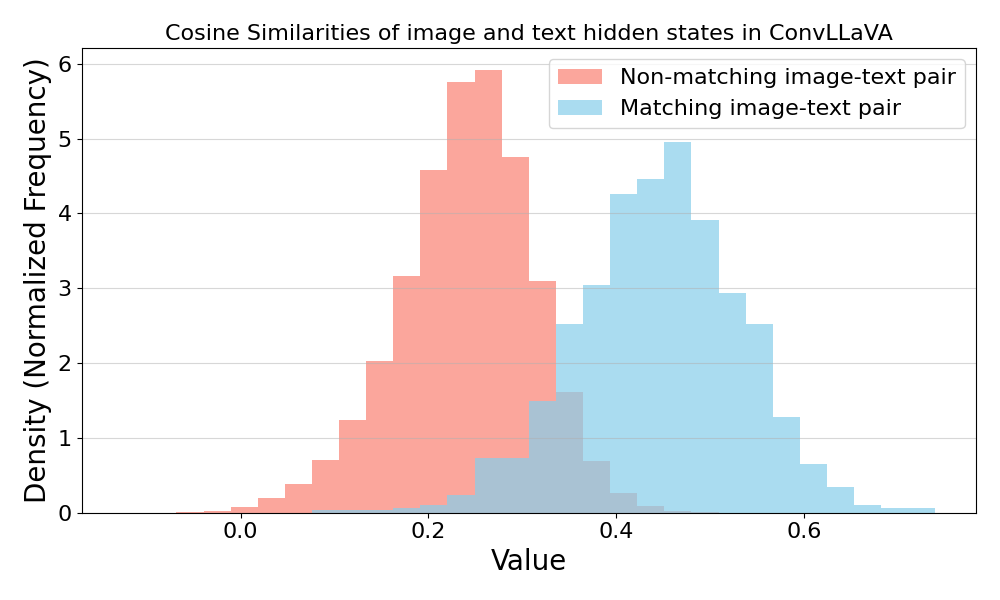}
        \caption{Similarity for ConvLLaVA \cite{ge2024convllava}}
    \end{subfigure}%
    \begin{subfigure}[t]{0.5\textwidth}
        \centering
        \includegraphics[width=\linewidth]{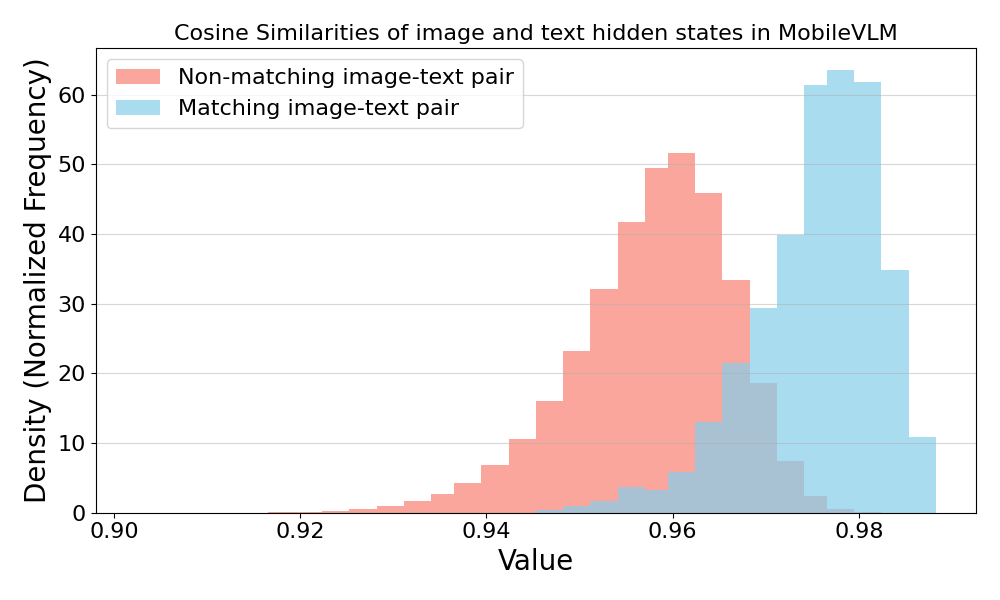}
        \caption{Similarity for MobileVLM-v2 \cite{chu2024mobilevlm}.}
    \end{subfigure}
    \caption{\textbf{Similarity analysis.} Cosine similarly between the hidden states of the images and text tokens of the last self-attention layer in existing efficient architectures. The similarities overlap for non-matching pairs 
    indicating a gap in the alignment of visual signal with text.}
    \label{fig:sim_hist}
\end{figure}

\section{Related Work}
\textbf{Efficient VLMs.} Vision–language models (VLMs) combine a visual encoder with a large language model to support multi-modal reasoning. 
Recent research has explored making VLMs more efficient and compact without sacrificing performance by using compressed image embedding and a smaller-sized language model. 
ConvLLaVA \cite{ge2024convllava} swaps the standard ViT \cite{dosovitskiy2020image} for a ConvNeXt \cite{liu2022convnet} encoder, cutting the number of visual‑token for high‑resolution images. 
FastVLM \cite{vasu2024fastvlm} introduces a hybrid vision encoder that yields far fewer tokens and achieves a better speed‑accuracy trade‑off. 
MobileVLM \cite{chu2024mobilevlm} reaches real‑time speed on edge devices through extensive ablation of design choices. 
VL‑Mamba \cite{qiao2024vl} replaces the Transformer with linear‑time state‑space (SSM) layers, delivering near‑linear scaling in sequence length while retaining competitive accuracy. 
Mini‑Gemini \cite{li2024mini} adopts a dual‑encoder scheme (low‑res ViT plus high‑res ConvNeXt); the visual‑token budget stays fixed, and high‑resolution details are injected only when needed. 
Finally, TinyLLaVA \cite{zhou2024tinyllava}, AppVLM \cite{papoudakis2025appvlm}, and VILA \cite{lin2024vila} report further gains from better training recipes, thorough dataset curation, and deep understanding of pre‑training. In this work, we develop efficient VLM framework which can effectively utilize the visual information and localize the correct regions in an image for visually-grounded conversational multimodal model.

\textbf{Attention in VLMs.} Most modern VLMs fuse vision and language information through carefully designed attention layers for comprehensive reasoning \cite{bhattacharyya2023look,cobbe2021training}. 
Flamingo \cite{alayrac2022flamingo} encodes images with a Perceiver module and feeds them into the language model via gated cross‑attention. 
BLIP‑2 \cite{li2023blip} uses a Q‑Former that queries image features and hands a compact token set to the LLM, injecting visual clues at multiple points to the language model with relatively few new parameters. 
Recent studies also highlight limitations in existing attention patterns and propose remedies.
\citeauthor{zhang2025mllms} \cite{zhang2025mllms} shows that performance falls sharply when the target object is small; a training‑free, attention‑guided cropping strategy recovers most of the lost accuracy. 
\citeauthor{bhattacharyya2023look} \cite{bhattacharyya2023look} interleaves top‑down cross‑attention blocks amid the LLM’s self‑attention, grounding generation in fine‑grained video frames. \citeauthor{kang2025see} \cite{kang2025see} observes biased attention toward irrelevant visual tokens and introduces visual‑attention‑sink suppression to redistribute focus and boost accuracy. Inspired by \citeauthor{bhattacharyya2023look} \cite{bhattacharyya2023look}, we introduce interleaved cross-attention layers into the self-attention layers of the LLM.

\textbf{Hallucination in VLMs.} Hallucination is well-known chronic problem in LVM. 
Misalignment between visual evidence and textual generation, especially in cluttered scenes, often drives such errors. 
Several benchmarks reveal that some VLMs perform similarly with or without visual input, implying that the language model may ignore image cues\cite{goyal2017making,kumar-etal-2024-vision,li2023evaluating,apollo,zhang2024debiasing}
To enhanced alignment, EMMA\cite{ghazanfari2024emma} balances structural and hierarchical representations, reducing hallucinated objects and sharpening visual grounding. 
Modular attribution studies find that multi‑head attention poses higher hallucination risk than MLP blocks; disabling “hallucination heads” yields simple yet effective mitigation \cite{yangmitigating}. 
New evaluation suites now include explicit hallucination tests \cite{guan2024hallusionbench,LiDZWZW23,tong2024cambrian}. 
For instance, the Multi‑Object Hallucination \cite{chen2024multi} dataset probes scenarios where models overlook vivid visual clues; its curated corner cases trace errors to language bias and skewed object distributions.

\section{Vision in Vision-Language Models}
\label{sec:method}

In this section, we will first investigate the underlying cause of object hallucination and limitations in processing the visual information in efficient VLMs. 
Based on our findings, we propose a framework to mitigate object hallucination in an efficient formulation. 

\subsection{Attention analysis in VLMs}
To understand the underlying causes of object hallucination and the tendency of VLMs--which are built on existing LLM backbones--to underutilize visual features, we analyze the semantic alignment between hidden states derived from their image and text modalities. Specifically, we compute the cosine similarity between the final hidden states of image tokens (or their projected representations) and text tokens, as visualized in Fig.~\ref{fig:sim_hist}. Our analysis considers two distinct concatenation-based models: ConvLLaVA, which pairs a convolutional vision backbone with a LLaMA-7B language model, and MobileVLM-v2, which utilizes a CLIP ViT-L/14 vision encoder with a LLaMA-1.4B. 

For both ConvLLaVA and MobileVLM-v2, a critical observation is the significant overlap in similarity score distributions between matching (ground-truth image-text pairs) and non-matching (randomly paired images and texts from the batch/dataset) examples. This suggests that these architectures systematically struggle to distinguish semantically coherent visual-textual pairs from incoherent ones in their hidden representations.

For ConvLLaVA (Fig.~\ref{fig:sim_hist}a), although the similarity scores for non-matching pairs are appropriately skewed towards lower values ($\sim 0.2-0.3$)--demonstrating some discriminative ability--the distribution for matching pairs is disappointingly centered around a modest $\sim 0.5$. Ideally, correctly matched pairs should exhibit a distribution strongly skewed towards higher scores (e.g., $>0.8$), signifying robust alignment between visual concepts and textual descriptions. This current observation implies that even when an image and text are semantically related, their respective hidden states are not achieving the desired close alignment in the shared embedding space.

MobileVLM-V2 (Fig.~\ref{fig:sim_hist}b) exhibits indiscriminately high similarity scores for both matching and non-matching pairs, with both distributions peaking at very high values (e.g., $\sim 0.96-0.98$). This consistent high similarity, irrespective of actual image-text semantic relevance, suggests a critical limitation in its ability to capture meaningful underlying multimodal semantic information.

This behavior is a strong indicator for object hallucination in VLMs and the dependence of the generation process on the strong language priors. Indeed, if non-matching pairs consistently achieve high similarity scores, it implies that visual features are failing to sufficiently constrain the LLM. Consequently, generation becomes unanchored from the visual input, driven instead by the LLM's internal biases or textual context, which leads to both object hallucination and a tendency to disregard specific visual details.
To mitigate object hallucination in efficient VLMs, our work introduces cross-attention layers whose attention weights are distilled from the Segmentation Anything Model (SAM), thereby better grounding the pretrained LLM in visual information.
\begin{figure}
    \centering
    \includegraphics[width=\linewidth]{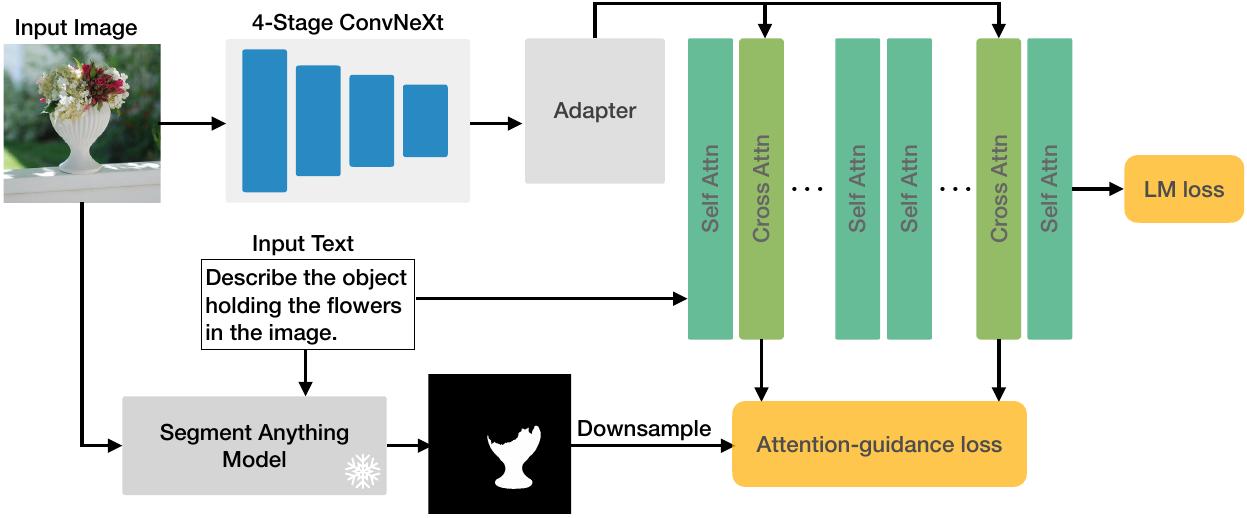}
    \caption{\textbf{Overall architecture of our attention-guided efficient vision language model.} During training, given the input image and the associated instruction, we perform knowledge distillation from SAM by explicitly aligning the language-conditioned masks with the cross-attention weights of our modified LLaMA-1B backbone.}
    \label{fig:archi}
    \vspace{-10pt}
\end{figure}

\subsection{Attention guided Efficient VLM Approach}
We present AGE-VLM, an efficient multimodal model that seamlessly integrates visual features with a language model architecture. AGE-VLM employs a ConvNext vision encoder and the LLaMA-1B decoder-only language model. 
The vision features are modulated by text tokens through cross-attention layers which are explicilty guided by distilling knowledge from SAM (\emph{c.f.}~Fig.~\ref{fig:archi}).

\subsubsection{Efficient Vision-Language Architecture}
\paragraph{Efficient vision backbone.} 

Similar to prior VLMs employing convolutional backbones, we utilize a ConvNeXT to extract visual features. Convolutional networks advantageously process higher-resolution images with fewer visual tokens compared to ViTs. Given an input image $I$ of spatial resolution $H \times W$, the ConvNeXT backbone processes it through multiple convolutional stages\footnote{Not to be confused with VLM training stages.}. We extract the spatial feature map $I'$ from the output of the fourth stage, which retains spatial information crucial for detailed visual understanding. This map $I'$ is then flattened and projected by two linear layers into a sequence of $h \times w$ visual tokens, each with dimension $d$ to match our language model's embedding dimension.

\paragraph{Efficient LLaMA-1B backbone.}
We employ LLaMA-1B as our language backbone, selected for its relatively small size, making it suitable for resource-constrained scenarios. The model processes tokenized text sequences. During training, most of LLaMA's parameters--specifically its self-attention and feed-forward network (FFN) weights--are kept frozen to preserve its powerful language priors and reduce training costs.

\paragraph{Interleaved cross-attention layers.}
To directly integrate visual information into the language model, we introduce cross-attention mechanisms within the LLaMA architecture. Instead of a simple prefix or concatenation approach, we interleave lightweight cross-attention modules into specific LLaMA decoder blocks. For LLaMA-1B, which has 16 decoder layers, these cross-attention modules are strategically inserted.
A standard LLaMA decoder layer $i$ typically processes input hidden states $H_{i-1}$ as follows:
\begin{align}
H_i &= \textsc{SelfAttention}(\textsc{Layernorm}(H_{i-1})), \nonumber\\
H_i &= H_i + H_{i-1}, \label{eq:sa_residual} \nonumber\\
H_i &= H_i + \textsc{MLP}(\textsc{Layernorm}(H_i)). 
\end{align}

We modify select layers--specifically those indexed 2, 7, 12, and 17 in LLaMA-1B--by inserting a cross-attention module after the standard self-attention sub-layer. This inserted cross-attention module takes the output of the self-attention sub-layer, $H_i$, and the visual features $I'$ (extracted by the ConvNeXT encoder and transformed by an adapter to match the LLM's hidden state dimension) as input. Within this cross-attention module, the hidden states from the self-attention sub-layer serve as queries $Q=W_q(H_i)$, while the transformed visual features serve as keys $K=W_k(I')$ and values $V = W_v(I')$. The operation is then:
\begin{align}
H_{\textsc{CA}} &= \textsc{CrossAttention}(Q, K, V), \nonumber\\
H_{\textsc{CA}} &= H_i + H_{\textsc{CA}},  \nonumber\\
H_{\textsc{CA}} &= H_{\textsc{CA}} + \textsc{MLP}(\textsc{Layernorm}(H_{\textsc{CA}})). 
\end{align}
This interleaved structure allows the model to dynamically ground textual concepts in visual features at multiple semantic levels within the LLM. The weights $W_q, W_k, W_v$ and the parameters within the CrossAttention blocks with multi-head attention are trainable.

\subsubsection{Spatial Grounding Distillation using SAM}
To account for the lack of spatial localization in VLMs optimized with the cross-entropy loss for next token prediction, we perform knowledge distillation from the Segment Anything Model in the cross-attention layers our model. 
For this, during pretraining stage, we take the 77K image-caption pairs corresponding to $\sim10$ percent of the pretraining data of Cambrian 2.5M. 
Analogously, during fine-tuning, we take the 150K image-instruction or image-question pairs from the Cambrian 10M. 
Using these language queries, we obtain the language-grounded masks for the images using SAM.

Given the image $I$ of spatial resolution $H \times W$, and the text prompt query $t_q$ (ignoring the special tokens) we obtain the mask $M \in \{0,1\}^{H \times W}$. 
The mask is then downsampled to match the vision feature encoder's spatial resolution $h \times w$ yielding  $M' \in \{0,1\}^{h \times w}$.
Given the attention weights $A_l$, output of softmax in cross-attention layer corresponding to layer $l$, where $l \in {l_1,\ldots,l_n}$ and $n$ is the number of cross-attention layers, the attention weights for the query token $t_q$ are averaged across all the heads in the attention layer and are reshaped to $h \times w$ yielding $A_l^q$ . 
Consider an example of a text prompt with 10 tokens and 576 image tokens, the cross-attention layers with 32 heads would ouput the attention weights $A_l$ of size $32 \times 10 \times 576$. These weights are averaged along the first two dimensions providing 576 dimensional $A_l^q$.
These attention weights  are then normalized to obtain a attention distribution $P_l^q$, 
We perform distillation using the dice loss to localize the attention maps on the region represented by the mask,
\begin{equation}
     \mathcal{L}_{g} = -\log \left[  \frac{2. \langle\text{vec}(M'). \text{vec}(P_l^q)\rangle }{ \sum_{i,j}M'_{i,j} + \sum_{i,j}P_l^q} \right]
     \label{eq:SAM}
\end{equation}
Here, vec(.) flattens the input to a 1-d representation. The advantage of dice loss is that it directly optimizes for the overlap between predicted and ground truth masks accounting for sparse regions of interest which are otherwise difficult to optimize using binary cross-entropy loss. This loss is applied to all the cross-attention layers to modulate the visual features with text tokens.

The overall training objective is the sum of the standard $\mathcal{L}_{LM}$ -- the causal language modeling (next-token prediction) loss computed using standard cross-entropy on the entire dataset and $\mathcal{L}_{g}$--the loss of distillation calculated on the subset with the SAM grounding masks.
The two loss are trained with equal weights.

\begin{table*}
\centering
\scriptsize
\caption{Training stages of our Attention-guided Efficient Vision Language Model.}
  \begin{tabularx}{\textwidth}{@{}Xcccccc@{}}
  \toprule
  & \multicolumn{4}{c}{Model} &  \multicolumn{2}{c}{Training Loss}\\
   \cmidrule(lr){2-5} \cmidrule(lr){6-7}
  Training Stage & Vision Encoder & Adapter & LLM(CA) &  LLM(SA) & LM Loss & Guidance Loss\\
  \hline
  Stage 1 & \textcolor{freezingblue}{\faSnowflake{}} & \textcolor{burningred}{\faFire{}} & \textcolor{burningred}{\faFire{}} & \textcolor{freezingblue}{\faSnowflake{}} & \checkmark & \XSolidBrush \\
   Stage 2 & \textcolor{burningred}{\faFire{}} &  \textcolor{burningred}{\faFire{}}& \textcolor{burningred}{\faFire{}} &\textcolor{freezingblue}{\faSnowflake{}}  & \checkmark & \XSolidBrush  \\
   Stage 3 &  \textcolor{burningred}{\faFire{}} &  \textcolor{burningred}{\faFire{}}& \textcolor{burningred}{\faFire{}} &\textcolor{freezingblue}{\faSnowflake{}}  & \checkmark & \checkmark\\
   Stage 4 (AGE-VLM) &  \textcolor{burningred}{\faFire{}} &  \textcolor{burningred}{\faFire{}}& \textcolor{burningred}{\faFire{}} &  \textcolor{burningred}{\faFire{}} &\checkmark & \XSolidBrush \\
   Stage 4 (AGE-VLM-LM) &  \textcolor{burningred}{\faFire{}} &  \textcolor{burningred}{\faFire{}}& \textcolor{burningred}{\faFire{}} &  \textcolor{burningred}{\faFire{}} &\checkmark &\checkmark \\ 
   \bottomrule
  \end{tabularx}
  \label{table:training_stages}
\end{table*}
\subsubsection{Training Stages}
Our model training, in Table~\ref{table:training_stages}, proceeds through four distinct stages. The first three stages comprise a comprehensive pre-training phase aimed at effectively aligning visual features with the textual representations of our lightweight 1B parameter LLM. A primary objective throughout this pre-training is to instill the LLM with visual capabilities while preserving its inherent language proficiency.

\paragraph{Stage1: Initial vision-language alignment.}
The first stage establishes a foundational mapping between modalities by aligning visual features (processed through an adapter) with the LLM's textual representations. We achieve this alignment using newly integrated cross-attention layers, with training guided exclusively by the LLM's inherent language modeling objective (e.g., next-token prediction). This provides strong initial weights for the adapter and cross-attention modules, teaching them to map visual information into the LLM's embedding space. This methodology, utilizing image-caption pairs from the Cambrian 2.5M dataset, is analogous to the initial pre-training phase of standard VLMs. 

\paragraph{Stage 2: Vision encoder adaptation.} In the second stage, we unfreeze and fine-tune the final block of the ConvNeXT vision encoder, training it jointly with the adapter and cross-attention layers. This approach is motivated by prior work demonstrating that adapting pre-trained ConvNeXt models from their original resolution (e.g., 384$\times$384) to higher resolutions (e.g., 768$\times$768) enhances detailed visual understanding. Operating at this higher resolution, our ConvNeXT yields 576 visual tokens, comparable to a Vision Transformer (ViT) backbone at a 336$\times$336 resolution. This highlights ConvNeXT's greater token efficiency compared to common ViT-based VLMs. The Cambrian 2.5M dataset continues to provide image-caption pairs for the LLM loss in this stage.

\paragraph{Stage 3: Spatial grounding via knowledge distillation and alignment.} The third stage enhances visual grounding by incorporating knowledge distillation from the Segment Anything Model (SAM) to ensure generated responses are explicitly tied to relevant visual information. For approximately 10\% of the Cambrian 2.5M image-text pairs, SAM generates segmentation masks for key entities or concepts relevant to the image-text context. We then optimize our model's cross-attention weights to align with these SAM-generated masks using the objective defined in Eq.~\ref{eq:SAM}. This encourages the cross-attention mechanism to focus on pertinent image regions during visual processing, thereby improving spatial grounding. The LLM loss is computed using the Cambrian 2.5M dataset.

\paragraph{Stage 4: Visually-grounded instruction fine-tuning.}
The final stage consists of end-to-end instruction fine-tuning for the entire model. 
We consider two variations for training.
In the first setting, we follow \cite{tong2024cambrian} (AGE-VLM) and finetune the model without the attention-grounding loss. The key advantage of this is that the self-attention layers of the LLM are kept intact, allowing to efficiently integrate multimodal signal with the model retains its language capacity. 
In the second scenario (AGE-VLM-LM) visual grounding is maintained by concurrently applying the distillation loss (from Stage 3) and the primary LLM loss (next-token prediction for instruction following). For knowledge distillation, SAM is prompted with the instruction (typically a question) and its ground-truth answer. This guides SAM to generate segmentation masks for image regions most pertinent to that specific instruction-answer pair, and our model's attention is then distilled towards these masks. The LLM loss in this stage utilizes the full Cambrian 10M instruction-following dataset, while the grounding loss is applied to a 10\% subset thereof, reinforcing the model's focus on relevant visual evidence.
\begin{table*}
\centering
\scriptsize
\caption{{\bf Quantitative evaluation.} Comparison of our AGE-VLM with state-of-the-art efficient VLMs on vision-centric benchmarks.}
  \begin{tabularx}{\linewidth}{@{}Xcccccccccc@{}}
  \toprule
    Method  &\multicolumn{3}{c}{HallusionBench} & \multicolumn{2}{c}{OCRBench}  &  \multicolumn{2}{c}{CV-Bench} & {RWQA} & POPE  \\
    \cmidrule(lr){2-4}  \cmidrule(lr){5-6}  \cmidrule(lr){7-8}
      & aAcc& fAcc & qAcc & Scene Centric & Key Info. Extract.  & 2D & 3D &&  \\
    \midrule
    CA-Baseline & 40.38 & 13.87 & 11.21 & 148.00 & 51.00 &  0.62 & 0.50 & 0.47 & 85.11 &\\
    ConvLLaVA & 24.71 & 8.96 & 4.84 & 117.00 & 26.00 & 0.59 & \bfseries 0.57 & \bfseries 0.51 & 77.76 \\
    mobile-vlm-v2 & \bfseries 44.37 & 14.45 & \bfseries 11.65 & 101.00 & 2.00 & 0.31 & 0.40 & 0.28 & 84.30&\\
 \midrule
   AGE-VLM &  43.85 & \bfseries 15.32 &  11.21 & \bfseries 149.00 & \bfseries 59.00 &  0.61 &  0.52 & 0.48 &  \bfseries 87.34 & \\
    AGE-VLM-LM & 39.22 & 11.56 & 7.91 & 126.00 & 33.00 & \bfseries 0.66 & 0.46 &  \bfseries 0.51 & 85.18 & \\
     \bottomrule
  \end{tabularx}
 
  \label{table:coco_quant}
   \vspace{-0.5 cm}
\end{table*}

\section{Experiments}
To validate the effectiveness of our AGE-VLM in encoding and utilizing visual features to mitigate object hallucination, following \cite{tong2024cambrian} we perform extensive experiments on vision-centric tasks  for objection hallucination evaluation on Visual Question Answering (VQA) with human edited images on HallusionBench \cite{guan2024hallusionbench} and on the POPE \cite{LiDZWZW23} dataset. 
Additionally,we include OCRBench \cite{liu2024ocrbench} for scene-centric text-VQA and for key information extraction from the receipt images.
We further evaluate on CV-Bench and RealWorldQA \cite{realworldqa}  to uncover multimodal capabilities in 2D tasks, i.e., spatial relationships or object count and for 3D tasks such as depth and relative distances.

\paragraph{Training setup.} 
As discussed in Sec.~\ref{sec:method}, we perform the four stage training of the model. For pretraining (stages 1, 2 and 3) we use Cambrian2.5M dataset and perform instruction finetuning (stage 4) with the Cambrian10M dataset. 
We train our model on 8 Nvidia A100 GPUs with a batchsize of 16 per GPU. 
We adopt the pretrained LLaMA-1B and the 4-stage constrastively trained ConvNeXt as the base model for our attention-guided efficient VLM. 
We train our approach with two variants AGE-VLM which does not apply attention-guidance loss during finetuning (stage 4). This variant applies attention grounding only in the pretraining stage. The key advantage of this is that the self-attention layers are kept intact yielding a efficient way of introducing multimodal capability in the text-only LLM.
In another scenario AGE-VLM-LM, we also apply guidance loss and train the model with the stage 4 as outlined in Sec.~\ref{sec:method}.
\begin{table}[t]
\caption{Qualitative comparison of our approach againt Conv-LLaVA and MobileVLM-v2 on vision-centric task.}
\scriptsize
\begin{tabularx}{\linewidth}{@{}lXXX@{}}
\toprule
& \raisebox{-51pt}{\includegraphics[width=\linewidth,height = 2.5 cm]{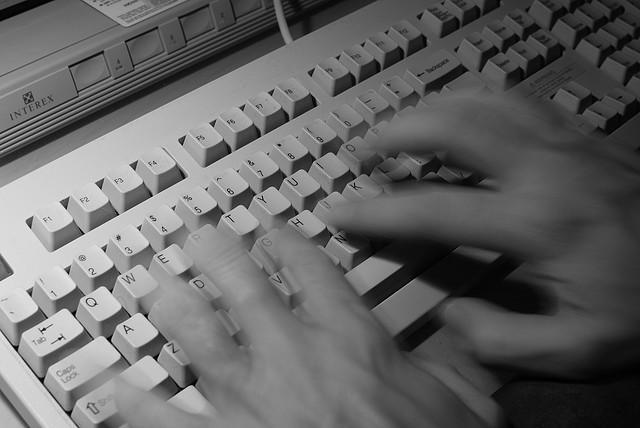}}
 & \raisebox{-51pt}{\includegraphics[width=\linewidth, height = 2.5 cm]{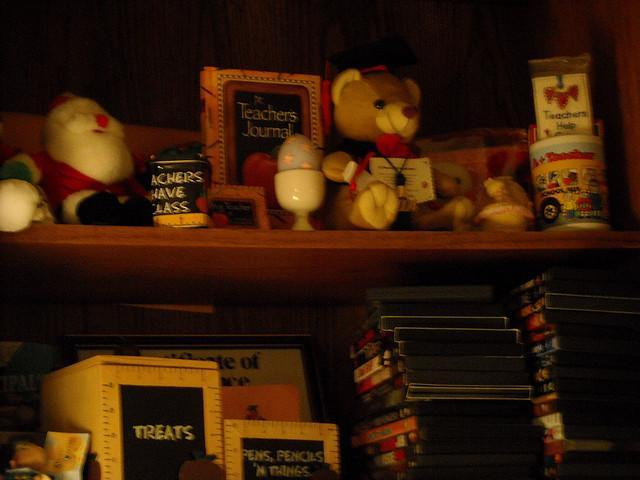}} &
  \raisebox{-51pt}{\includegraphics[width=\linewidth,height = 2.5 cm]{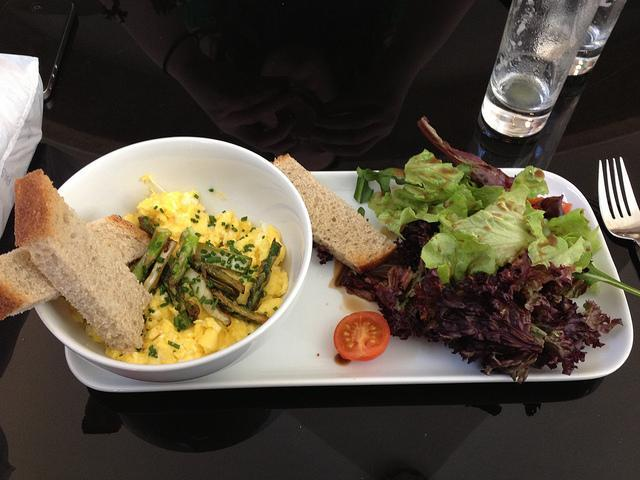}}
 \\
Prompt  &  Is there a laptop in the image? Please answer yes or no. & Is there an apple in the image? Please answer yes or no. & Is there a cup in the image? Please answer yes or no.\\
Groundtruth & yes & yes & yes\\
\midrule
Conv-LLaVA \cite{ge2024convllava} & 0 & 0 & no\\
\midrule
MobileVLM-v2 \cite{chu2024mobilevlm} &  No, there is no laptop in the image. The image features a person typing on a keyboard. & No, there is no apple in the image. The image features a shelf with various items, including books, a teddy bear, and a Christmas ornament. & No, there is no cup in the image. The image only shows a plate of food, including a salad and a bowl of soup.  \\
\midrule
AGE-VLM & Yes, there is a laptop in the image. It is partially visible in the background, partially obscured by the hands of the person typing on the keyboard. The laptop is positioned to the left side of the image, and it appears to be a standard model with a white exterior and black keys. &  Yes, there is an apple in the image. It is located on the top shelf, to the left of the teddy bear. The apple is red with a yellow stem and is placed among other items. & Yes, there is a cup in the image. It is located on the right side of the plate, partially obscured by the bread. The cup appears to be a clear glass, and it is filled with a transparent liquid, which could be water or another clear beverage.\\
 \bottomrule
\end{tabularx}
\label{table:Qualitative}
\vspace{-0.5 cm}
\end{table}
\paragraph{Evaluation metrics.}
We evaluate our models on diverse benchmarks with each having a different metric to assess model performance. 
HallusionBench considers \textit{aAcc}: the overall accuracy of all atomic questions, 
\textit{qAcc}: the mean accuracy of unique questions as one question can be asked multiple times with different figures. A VLM correctly solved a unique question only if it succeeds in all <question, figure> pairs for this unique question.
\textit{fAcc}: the mean accuracy of all figures. One figure is associated with multiple questions, a VLM iscorrect on a figure only if it succeeds to solve all questions of this figure.
CV-Bench consists of multiple choice questions, the models however, sometimes do not output the option even though they generate the correct answer. To account for this, we evaluate the accuracy by employing Qwen-L for evaluation.
For OCRBench and RealWorldQA,  we report the accuracy on the Scene-centric and the key information extraction tasks. 

\begin{table}[t]
\scriptsize
\caption{\textbf{Attention visualization for different models.} Our method looks at the right regions based on the input image and the input text.}
\smallskip
\begin{tabularx}{\linewidth}{@{}XXXX@{}}
\toprule
Image & Conv-LLaVA \cite{ge2024convllava} & CA-baseline & Ours \\
\midrule
 \raisebox{-51pt}{\includegraphics[width=\linewidth]{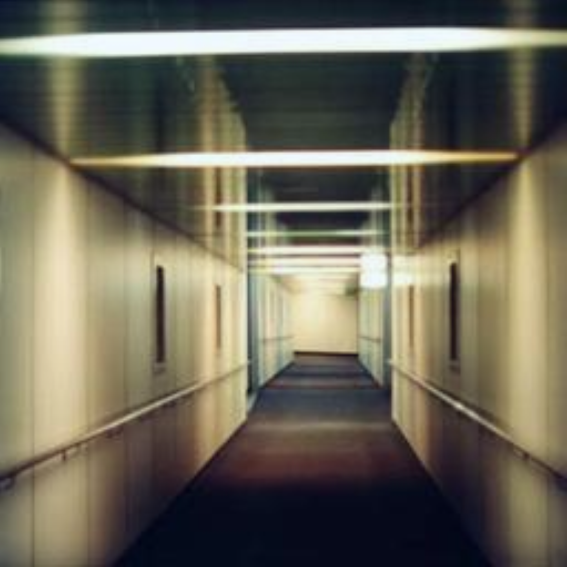}}
 & \raisebox{-51pt}{\includegraphics[width=\linewidth]{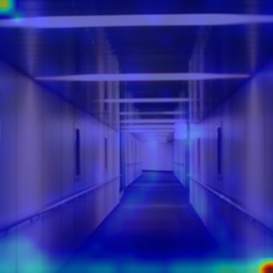}}
&\raisebox{-51pt}{\includegraphics[width=\linewidth]{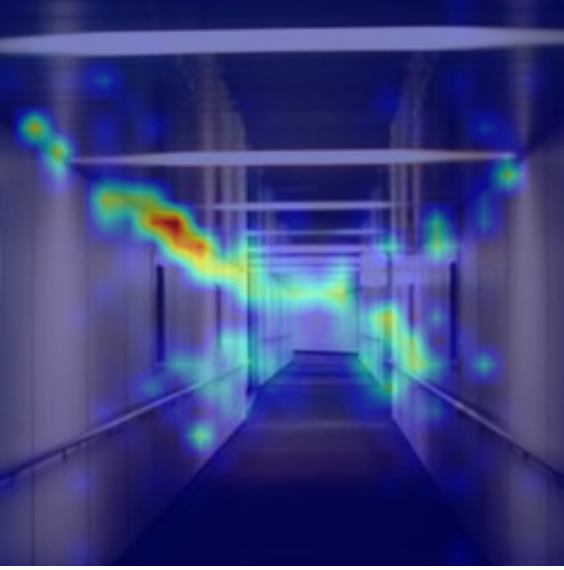}}

& \raisebox{-51pt}{\includegraphics[width=\linewidth]{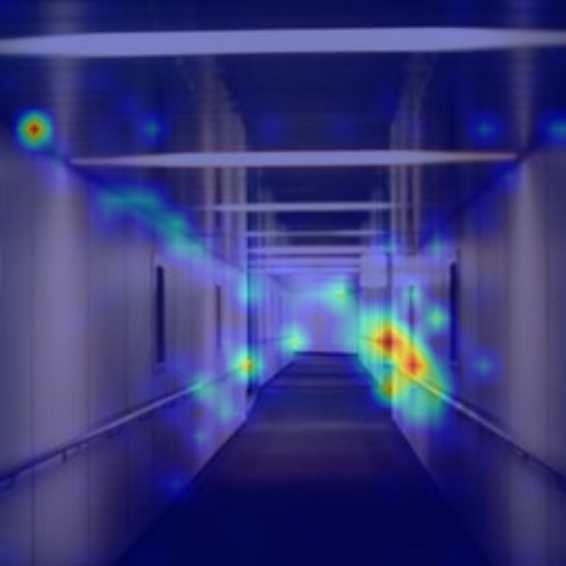}}\\
\multicolumn{4}{c}{How many handrails are in the image?}\\
\midrule
\raisebox{-51pt}{\includegraphics[width=\linewidth]{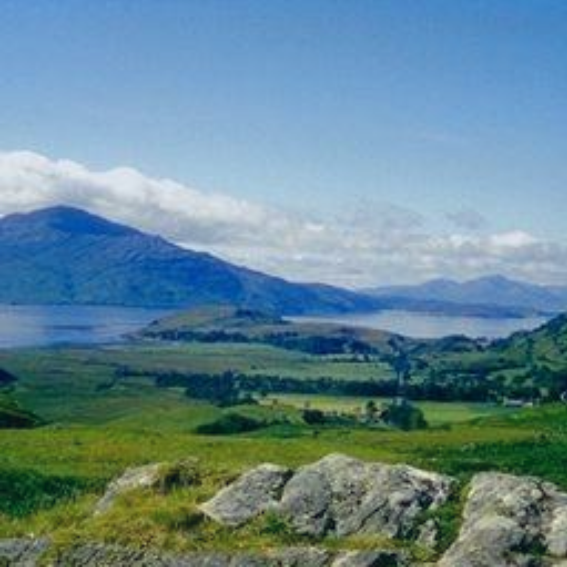}}
 & \raisebox{-51pt}{\includegraphics[width=\linewidth]{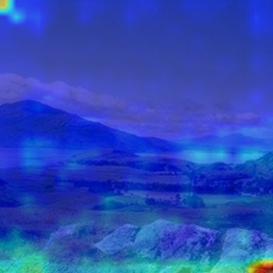}}
&\raisebox{-51pt}{\includegraphics[width=\linewidth]{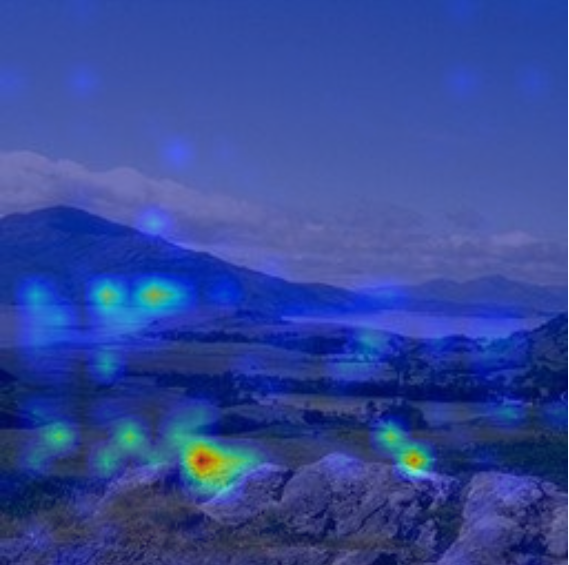}}

& \raisebox{-51pt}{\includegraphics[width=\linewidth]{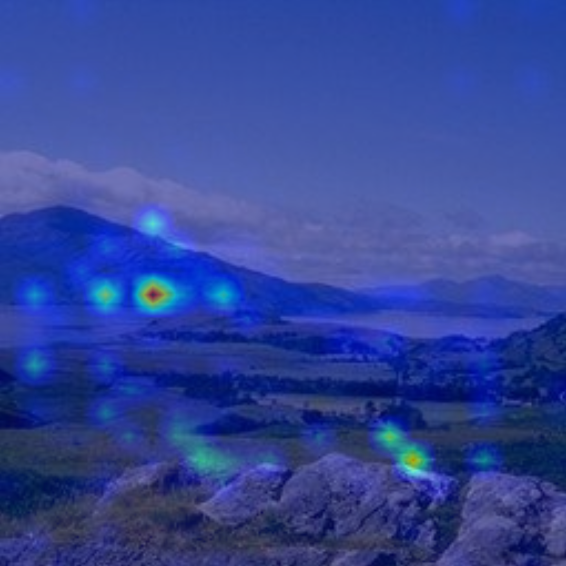}}\\
\multicolumn{4}{m{0.99\textwidth}}{Considering the relative positions of the river water and the stone in the image provided, where is the river water located with respect to the stone?}\\
\bottomrule
\end{tabularx}
\label{table:attention_vis}
\vspace{-0.5 cm}
\end{table}

\paragraph{Prior-art and baseline.}
We compare our approach against ConvLLaVA, MobileVLM-v2 and CA-baseline. 
ConvLLaVA also extracts vision features from ConvNeXt which are input to Vicuna-7B. MobileVLM-v2 with 1.7B parameters is based on CLIP-ViT with their MobileLLaMA, a downsized version of LLaMA.
Both the models concatenate the vision tokens to the language tokens which are input to their respective LLMs optimized with the LM loss. 
We also include CA-baseline which has all the elements of our approach except for spatial distillation which attention guidance. That is in this variant the cross-attention layers and the self-attention layers are trained using only the LM loss. 

\paragraph{Quantitative results.} In Tab.~\ref{table:coco_quant} we compare our approach to the prior-art and the baselines on efficient VLMs on different vision-centric benchmarks. 
We observe that on challenging datasets such as CV-bench our model outperforms prior work by a large margin.
Similar improvements are demonstrated on the OCRbench and the RealWorldQA  datasets. 
This highlights the enhanced vision processing ability of our approach.
Furthermore, we note that while our approach on HallusionBench yields better performance than ConvLLaVA, it is comparable to that of MobileVLM-v2.
This can be attributed to fact that the attention signal from SAM cannot text information in mathematical charts or figures.
Notably, our AGE-VLM variant trained in an efficient manner with self-attention layers intact (not trained with the grounding loss) consistently outperforms prior art with seamless integration of visual information with just 1.2B parameters.

\paragraph{Qualitative Results.}
In Tab.~\ref{table:Qualitative} we present the qualitative comparison of our AGE-VLM approach against Conv-LLaVA and MobileVLM-v2, the prior art on efficient VLMs on vision-centric question answering task for challenging question and image pairs. 
Even though Conv-LLaVA answers incorrectly, it adheres to the instruction, answering with 0 or no.  
The responses generated by MobileVLM-v2 are not well grounded in the image as is evident from the explanation that follows the answer.
For example, in column 3, MobileVLM-v2 incorrectly generates ``soup'' as the item in the image.
In contrast, our approach not only follows the instruction but can also generate the response grounded in the image information. 
For a challenging case in column 2, our approach correctly localizes the location of apple in terms of the spatial relationship with other objects in the image and provides the correct response. 
We demonstrate the localization capabilities of our approach in Tab. ~\ref{table:attention_vis}. 
We visualize the attention weights of the first self-attention layer for Conv-LLaVA and the first cross-attention layer for the CA-baseline without attention guidance and our approach with attention guidance. 
As shown, given the image and the associated prompt, 
the Conv-LLaVA approach does not have any implicit grounding in the sefl-attention layer. 
The CA-baseline does have implicit grounding capacity but it incorrectly localizes the target visual concepts from the prompt. 
Our approach, on the other hand,  localizes the correct regions, e.g., handrails in row~1, and the river and stone in row~2.

\section{Conclusion}
We introduced AGE-VLM, an efficient VLM designed to mitigate object hallucination. Our findings demonstrate that distilling knowledge from the SAM to guide attention mechanisms significantly enhances the visual grounding of VLMs. Extensive experiments show AGE-VLM achieves performance that is markedly improved or comparable to existing efficient VLMs on various vision-centric benchmarks. We believe this approach will stimulate further research into efficiently aligning vision and text modalities within the hidden states of pretrained LLMs with minimal overhead.
\paragraph{Limitations and Broader Impact.} 
While this paper has focused on the training recipe for distilling knowledge from SAM into vision-language models, our approach does not explore scaling of the distillation data or consider distilling optical flow or object tracking into the hidden states of VLMs.
While our work addresses hallucination, it is far from perfect and can produce biased or factually incorrect content.
With efficient VLMs as proposed in this work gaining traction, they will be widely accessable and should therefore be used with caution as their incorrect responses can cause physical harm such when using self-diagnosis with consulting medical experts.

\bibliographystyle{plainnat}
{\small
\bibliography{2025-mahajan-neurips/neurips_2025}

@article{grattafiori2024llama,
  title={The llama 3 herd of models},
  author={Grattafiori, Aaron and Dubey, Abhimanyu and Jauhri, Abhinav and Pandey, Abhinav and Kadian, Abhishek and Al-Dahle, Ahmad and Letman, Aiesha and Mathur, Akhil and Schelten, Alan and Vaughan, Alex and others},
  journal={arXiv preprint arXiv:2407.21783},
  year={2024}
}

@article{alayrac2022flamingo,
  title={Flamingo: a visual language model for few-shot learning},
  author={Alayrac, Jean-Baptiste and Donahue, Jeff and Luc, Pauline and Miech, Antoine and Barr, Iain and Hasson, Yana and Lenc, Karel and Mensch, Arthur and Millican, Katherine and Reynolds, Malcolm and others},
  journal={Advances in Neural Information Processing Systems},
  volume={35},
  pages={23716--23736},
  year={2022}
}

@inproceedings{LiDZWZW23,
  author       = {Yifan Li and
                  Yifan Du and
                  Kun Zhou and
                  Jinpeng Wang and
                  Wayne Xin Zhao and
                  Ji{-}Rong Wen},
  editor       = {Houda Bouamor and
                  Juan Pino and
                  Kalika Bali},
  title        = {Evaluating Object Hallucination in Large Vision-Language Models},
  booktitle    = {Proceedings of the Conference on Empirical Methods in Natural Language Processing},
  pages        = {292--305},
  publisher    = {Association for Computational Linguistics},
  year         = {2023}
}

@article{tong2024cambrian,
  title={Cambrian-1: A fully open, vision-centric exploration of multimodal llms},
  author={Tong, Peter and Brown, Ellis and Wu, Penghao and Woo, Sanghyun and IYER, Adithya Jairam Vedagiri and Akula, Sai Charitha and Yang, Shusheng and Yang, Jihan and Middepogu, Manoj and Wang, Ziteng and others},
  journal={Advances in Neural Information Processing Systems},
  volume={37},
  pages={87310--87356},
  year={2024}
}

@inproceedings{guan2024hallusionbench,
  title={Hallusionbench: an advanced diagnostic suite for entangled language hallucination and visual illusion in large vision-language models},
  author={Guan, Tianrui and Liu, Fuxiao and Wu, Xiyang and Xian, Ruiqi and Li, Zongxia and Liu, Xiaoyu and Wang, Xijun and Chen, Lichang and Huang, Furong and Yacoob, Yaser and others},
  booktitle={Proceedings of the IEEE/CVF Conference on Computer Vision and Pattern Recognition},
  pages={14375--14385},
  year={2024}
}

@inproceedings{chu2024mobilevlm,
  author       = {Qinzhuo Wu and
                  Weikai Xu and
                  Wei Liu and
                  Tao Tan and
                  Jianfeng Liu and
                  Ang Li and
                  Jian Luan and
                  Bin Wang and
                  Shuo Shang},
  title        = {MobileVLM: {A} Vision-Language Model for Better Intra- and Inter-UI
                  Understanding},
  booktitle    = {{EMNLP} (Findings)},
  pages        = {10231--10251},
  year         = {2024}
}

@article{ge2024convllava,
  title={Convllava: Hierarchical backbones as visual encoder for large multimodal models},
  author={Ge, Chunjiang and Cheng, Sijie and Wang, Ziming and Yuan, Jiale and Gao, Yuan and Song, Jun and Song, Shiji and Huang, Gao and Zheng, Bo},
  journal={arXiv preprint arXiv:2405.15738},
  year={2024}
}

@misc{realworldqa,
  title={Grok-1.5 vision preview},
  author={x.ai},
  year={2024}
}

@article{liu2024ocrbench,
  title={OCRBench: on the hidden mystery of OCR in large multimodal models},
  author={Liu, Yuliang and Li, Zhang and Huang, Mingxin and Yang, Biao and Yu, Wenwen and Li, Chunyuan and Yin, Xu-Cheng and Liu, Cheng-Lin and Jin, Lianwen and Bai, Xiang},
  journal={Science China Information Sciences},
  volume={67},
  number={12},
  year={2024}
}

@article{liu2023visual,
  title={Visual instruction tuning},
  author={Liu, Haotian and Li, Chunyuan and Wu, Qingyang and Lee, Yong Jae},
  journal={Advances in Neural Information Processing Systems},
  volume={36},
  pages={34892--34916},
  year={2023}
}

@article{wang2024qwen2,
  title={Qwen2-vl: Enhancing vision-language model's perception of the world at any resolution},
  author={Wang, Peng and Bai, Shuai and Tan, Sinan and Wang, Shijie and Fan, Zhihao and Bai, Jinze and Chen, Keqin and Liu, Xuejing and Wang, Jialin and Ge, Wenbin and others},
  journal={arXiv preprint arXiv:2409.12191},
  year={2024}
}

@inproceedings{radford2021learning,
  title={Learning transferable visual models from natural language supervision},
  author={Radford, Alec and Kim, Jong Wook and Hallacy, Chris and Ramesh, Aditya and Goh, Gabriel and Agarwal, Sandhini and Sastry, Girish and Askell, Amanda and Mishkin, Pamela and Clark, Jack and others},
  booktitle={International Conference on Machine Learning},
  pages={8748--8763},
  year={2021}
}

@article{zhu2023minigpt,
  title={Minigpt-4: Enhancing vision-language understanding with advanced large language models},
  author={Zhu, Deyao and Chen, Jun and Shen, Xiaoqian and Li, Xiang and Elhoseiny, Mohamed},
  journal={arXiv preprint arXiv:2304.10592},
  year={2023}
}

@article{vasu2024fastvlm,
  title={FastVLM: Efficient Vision Encoding for Vision Language Models},
  author={Vasu, Pavan Kumar Anasosalu and Faghri, Fartash and Li, Chun-Liang and Koc, Cem and True, Nate and Antony, Albert and Santhanam, Gokul and Gabriel, James and Grasch, Peter and Tuzel, Oncel and others},
  journal={arXiv preprint arXiv:2412.13303},
  year={2024}
}

@article{cobbe2021training,
  title={Training verifiers to solve math word problems},
  author={Cobbe, Karl and Kosaraju, Vineet and Bavarian, Mohammad and Chen, Mark and Jun, Heewoo and Kaiser, Lukasz and Plappert, Matthias and Tworek, Jerry and Hilton, Jacob and Nakano, Reiichiro and others},
  journal={arXiv preprint arXiv:2110.14168},
  year={2021}
}

@article{le2022coderl,
  title={Coderl: Mastering code generation through pretrained models and deep reinforcement learning},
  author={Le, Hung and Wang, Yue and Gotmare, Akhilesh Deepak and Savarese, Silvio and Hoi, Steven Chu Hong},
  journal={Advances in Neural Information Processing Systems},
  volume={35},
  pages={21314--21328},
  year={2022}
}

@article{chowdhery2023palm,
  title={Palm: Scaling language modeling with pathways},
  author={Chowdhery, Aakanksha and Narang, Sharan and Devlin, Jacob and Bosma, Maarten and Mishra, Gaurav and Roberts, Adam and Barham, Paul and Chung, Hyung Won and Sutton, Charles and Gehrmann, Sebastian and others},
  journal={Journal of Machine Learning Research},
  volume={24},
  number={240},
  pages={1--113},
  year={2023}
}

@article{achiam2023gpt,
  title={Gpt-4 technical report},
  author={Achiam, Josh and Adler, Steven and Agarwal, Sandhini and Ahmad, Lama and Akkaya, Ilge and Aleman, Florencia Leoni and Almeida, Diogo and Altenschmidt, Janko and Altman, Sam and Anadkat, Shyamal and others},
  journal={arXiv preprint arXiv:2303.08774},
  year={2023}
}

@inproceedings{tong2024eyes,
  title={Eyes wide shut? exploring the visual shortcomings of multimodal llms},
  author={Tong, Shengbang and Liu, Zhuang and Zhai, Yuexiang and Ma, Yi and LeCun, Yann and Xie, Saining},
  booktitle={Proceedings of the IEEE/CVF Conference on Computer Vision and Pattern Recognition},
  pages={9568--9578},
  year={2024}
}

@inproceedings{rahmanzadehgervi2024vision,
  title={Vision language models are blind},
  author={Rahmanzadehgervi, Pooyan and Bolton, Logan and Taesiri, Mohammad Reza and Nguyen, Anh Totti},
  booktitle={Proceedings of the Asian Conference on Computer Vision},
  pages={18--34},
  year={2024}
}

@inproceedings{yue2024mmmu,
  title={Mmmu: A massive multi-discipline multimodal understanding and reasoning benchmark for expert agi},
  author={Yue, Xiang and Ni, Yuansheng and Zhang, Kai and Zheng, Tianyu and Liu, Ruoqi and Zhang, Ge and Stevens, Samuel and Jiang, Dongfu and Ren, Weiming and Sun, Yuxuan and others},
  booktitle={Proceedings of the IEEE/CVF Conference on Computer Vision and Pattern Recognition},
  pages={9556--9567},
  year={2024}
}

@article{lu2023mathvista,
  title={Mathvista: Evaluating mathematical reasoning of foundation models in visual contexts},
  author={Lu, Pan and Bansal, Hritik and Xia, Tony and Liu, Jiacheng and Li, Chunyuan and Hajishirzi, Hannaneh and Cheng, Hao and Chang, Kai-Wei and Galley, Michel and Gao, Jianfeng},
  journal={arXiv preprint arXiv:2310.02255},
  year={2023}
}

@article{hiippala2021ai2d,
  title={AI2D-RST: a multimodal corpus of 1000 primary school science diagrams},
  author={Hiippala, Tuomo and Alikhani, Malihe and Haverinen, Jonas and Kalliokoski, Timo and Logacheva, Evanfiya and Orekhova, Serafina and Tuomainen, Aino and Stone, Matthew and Bateman, John A},
  journal={Language Resources and Evaluation},
  volume={55},
  pages={661--688},
  year={2021}
}

@article{masry2025alignvlm,
  title={Alignvlm: Bridging vision and language latent spaces for multimodal understanding},
  author={Masry, Ahmed and Rodriguez, Juan A and Zhang, Tianyu and Wang, Suyuchen and Wang, Chao and Feizi, Aarash and Suresh, Akshay Kalkunte and Puri, Abhay and Jian, Xiangru and No{\"e}l, Pierre-Andr{\'e} and others},
  journal={arXiv preprint arXiv:2502.01341},
  year={2025}
}

@inproceedings{liu2022convnet,
  title={A convnet for the 2020s},
  author={Liu, Zhuang and Mao, Hanzi and Wu, Chao-Yuan and Feichtenhofer, Christoph and Darrell, Trevor and Xie, Saining},
  booktitle={Proceedings of the IEEE/CVF Conference on Computer Vision and Pattern Recognition},
  pages={11976--11986},
  year={2022}
}

@article{lin2023sphinx,
  title={Sphinx: The joint mixing of weights, tasks, and visual embeddings for multi-modal large language models},
  author={Lin, Ziyi and Liu, Chris and Zhang, Renrui and Gao, Peng and Qiu, Longtian and Xiao, Han and Qiu, Han and Lin, Chen and Shao, Wenqi and Chen, Keqin and others},
  journal={arXiv preprint arXiv:2311.07575},
  year={2023}
}

@article{oquab2023dinov2,
  author       = {Maxime Oquab and
                  Timoth{\'{e}}e Darcet and
                  Th{\'{e}}o Moutakanni and
                  Huy V. Vo and
                  Marc Szafraniec and
                  Vasil Khalidov and
                  Pierre Fernandez and
                  Daniel Haziza and
                  Francisco Massa and
                  Alaaeldin El{-}Nouby and
                  Mido Assran and
                  Nicolas Ballas and
                  Wojciech Galuba and
                  Russell Howes and
                  Po{-}Yao Huang and
                  Shang{-}Wen Li and
                  Ishan Misra and
                  Michael Rabbat and
                  Vasu Sharma and
                  Gabriel Synnaeve and
                  Hu Xu and
                  Herv{\'{e}} J{\'{e}}gou and
                  Julien Mairal and
                  Patrick Labatut and
                  Armand Joulin and
                  Piotr Bojanowski},
  title        = {DINOv2: Learning Robust Visual Features without Supervision},
  journal      = {Transactions on Machine Learning Research},
  volume       = {2024},
  year         = {2024}
}

@inproceedings{dosovitskiy2020image,
  author       = {Alexey Dosovitskiy and
                  Lucas Beyer and
                  Alexander Kolesnikov and
                  Dirk Weissenborn and
                  Xiaohua Zhai and
                  Thomas Unterthiner and
                  Mostafa Dehghani and
                  Matthias Minderer and
                  Georg Heigold and
                  Sylvain Gelly and
                  Jakob Uszkoreit and
                  Neil Houlsby},
  title        = {An Image is Worth 16x16 Words: Transformers for Image Recognition
                  at Scale},
  booktitle    = {International Conference on Learning Representations},
  year         = {2021}
}

@inproceedings{kirillov2023segment,
  title={Segment anything},
  author={Kirillov, Alexander and Mintun, Eric and Ravi, Nikhila and Mao, Hanzi and Rolland, Chloe and Gustafson, Laura and Xiao, Tete and Whitehead, Spencer and Berg, Alexander C and Lo, Wan-Yen and others},
  booktitle={Proceedings of the IEEE/CVF International Conference on Computer Vision},
  pages={4015--4026},
  year={2023}
}

@article{qiao2024vl,
  title={Vl-mamba: Exploring state space models for multimodal learning},
  author={Qiao, Yanyuan and Yu, Zheng and Guo, Longteng and Chen, Sihan and Zhao, Zijia and Sun, Mingzhen and Wu, Qi and Liu, Jing},
  journal={arXiv preprint arXiv:2403.13600},
  year={2024}
}

@article{li2024mini,
  title={Mini-gemini: Mining the potential of multi-modality vision language models},
  author={Li, Yanwei and Zhang, Yuechen and Wang, Chengyao and Zhong, Zhisheng and Chen, Yixin and Chu, Ruihang and Liu, Shaoteng and Jia, Jiaya},
  journal={arXiv preprint arXiv:2403.18814},
  year={2024}
}

@article{zhou2024tinyllava,
  title={Tinyllava: A framework of small-scale large multimodal models},
  author={Zhou, Baichuan and Hu, Ying and Weng, Xi and Jia, Junlong and Luo, Jie and Liu, Xien and Wu, Ji and Huang, Lei},
  journal={arXiv preprint arXiv:2402.14289},
  year={2024}
}

@article{papoudakis2025appvlm,
  title={AppVLM: A Lightweight Vision Language Model for Online App Control},
  author={Papoudakis, Georgios and Coste, Thomas and Wu, Zhihao and Hao, Jianye and Wang, Jun and Shao, Kun},
  journal={arXiv preprint arXiv:2502.06395},
  year={2025}
}

@inproceedings{lin2024vila,
  title={Vila: On pre-training for visual language models},
  author={Lin, Ji and Yin, Hongxu and Ping, Wei and Molchanov, Pavlo and Shoeybi, Mohammad and Han, Song},
  booktitle={Proceedings of the IEEE/CVF Conference on Computer Vision and Pattern Recognition},
  pages={26689--26699},
  year={2024}
}

@inproceedings{bhattacharyya2023look,
  author       = {Apratim Bhattacharyya and
                  Sunny Panchal and
                  Reza Pourreza and
                  Mingu Lee and
                  Pulkit Madan and
                  Roland Memisevic},
  title        = {Look, Remember and Reason: Grounded Reasoning in Videos with Language
                  Models},
  booktitle    = {International Conference on Learning Representations},
  year         = {2024}
}

@inproceedings{li2023blip,
  title={Blip-2: Bootstrapping language-image pre-training with frozen image encoders and large language models},
  author={Li, Junnan and Li, Dongxu and Savarese, Silvio and Hoi, Steven},
  booktitle={International Conference on Machine Learning},
  pages={19730--19742},
  year={2023},
  organization={PMLR}
}

@article{zhang2025mllms,
  title={MLLMs know where to look: Training-free perception of small visual details with multimodal LLMs},
  author={Zhang, Jiarui and Khayatkhoei, Mahyar and Chhikara, Prateek and Ilievski, Filip},
  journal={arXiv preprint arXiv:2502.17422},
  year={2025}
}

@article{ghazanfari2024emma,
  title={EMMA: Efficient Visual Alignment in Multi-Modal LLMs},
  author={Ghazanfari, Sara and Araujo, Alexandre and Krishnamurthy, Prashanth and Garg, Siddharth and Khorrami, Farshad},
  journal={arXiv preprint arXiv:2410.02080},
  year={2024}
}

@inproceedings{yangmitigating,
  title={Mitigating Hallucination in Large Vision-Language Models via Modular Attribution and Intervention},
  author={Yang, Tianyun and Li, Ziniu and Cao, Juan and Xu, Chang},
  booktitle={Adaptive Foundation Models: Evolving AI for Personalized and Efficient Learning},
  year = {2025}
}

@article{chen2024multi,
  title={Multi-object hallucination in vision language models},
  author={Chen, Xuweiyi and Ma, Ziqiao and Zhang, Xuejun and Xu, Sihan and Qian, Shengyi and Yang, Jianing and Fouhey, David and Chai, Joyce},
  journal={Advances in Neural Information Processing Systems},
  volume={37},
  pages={44393--44418},
  year={2024}
}

@inproceedings{
kang2025see,
title={See What You Are Told: Visual Attention Sink in Large Multimodal Models},
author={Seil Kang and Jinyeong Kim and Junhyeok Kim and Seong Jae Hwang},
booktitle={International Conference on Learning Representations},
year={2025}
}

@article{apollo,
    title={Apollo: An Exploration of Video Understanding in Large Multimodal Models},
    author={Orr Zohar and Xiaohan Wang and  Yann Dubois and Nikhil Mehta and Tong Xiao and Philippe Hansen-Estruch and Licheng Yu and Xiaofang Wang and Felix Juefei-Xu and Ning Zhang and Serena Yeung-Levy and Xide Xia},
    journal={arXiv preprint arXiv:2412.10360},
    year={2024}
}

@inproceedings{goyal2017making,
  title={Making the v in vqa matter: Elevating the role of image understanding in visual question answering},
  author={Goyal, Yash and Khot, Tejas and Summers-Stay, Douglas and Batra, Dhruv and Parikh, Devi},
  booktitle={Proceedings of the IEEE Conference on Computer Vision and Pattern Recognition},
  pages={6904--6913},
  year={2017}
}

@article{zhang2024debiasing,
  title={Debiasing multimodal large language models},
  author={Zhang, Yi-Fan and Yu, Weichen and Wen, Qingsong and Wang, Xue and Zhang, Zhang and Wang, Liang and Jin, Rong and Tan, Tieniu},
  journal={arXiv preprint arXiv:2403.05262},
  year={2024}
}

@inproceedings{
li2023evaluating,
title={Evaluating Object Hallucination in Large Vision-Language Models},
author={Yifan Li and Yifan Du and Kun Zhou and Jinpeng Wang and Xin Zhao and Ji-Rong Wen},
booktitle={Proceedings of the Conference on Empirical Methods in Natural Language Processing},
year={2023}
}

@inproceedings{kumar-etal-2024-vision,
    title = "Do Vision-Language Models Understand Compound Nouns?",
    author = "Kumar, Sonal  and
      Ghosh, Sreyan  and
      Sakshi, S  and
      Tyagi, Utkarsh  and
      Manocha, Dinesh",
    booktitle = "Proceedings of the  Conference of the North American Chapter of the Association for Computational Linguistics: Human Language Technologies (Volume 2: Short Papers)",
    year = "2024",
    pages = "519--527",
    
}

@article{ren2024grounded,
  title={Grounded sam: Assembling open-world models for diverse visual tasks},
  author={Ren, Tianhe and Liu, Shilong and Zeng, Ailing and Lin, Jing and Li, Kunchang and Cao, He and Chen, Jiayu and Huang, Xinyu and Chen, Yukang and Yan, Feng and others},
  journal={arXiv preprint arXiv:2401.14159},
  year={2024}
}

@inproceedings{0004WXDHL00Y24,
  author       = {Bin Xiao and
                  Haiping Wu and
                  Weijian Xu and
                  Xiyang Dai and
                  Houdong Hu and
                  Yumao Lu and
                  Michael Zeng and
                  Ce Liu and
                  Lu Yuan},
  title        = {Florence-2: Advancing a Unified Representation for a Variety of Vision
                  Tasks},
  booktitle    = {Proceedings of the {IEEE/CVF} Conference on Computer Vision and Pattern Recognition},
  pages        = {4818--4829},
  publisher    = {{IEEE}},
  year         = {2024}
}
}
\newpage

\appendix
\section*{Appendix}
We provide details on data collection for alignment guidance, additional details for training AGE-VLM, provide insights on further improvements with respect to the image data processing and include additional details on the evaluation benchmarks considered in the main paper. 
We also provide additional qualitative results. 
\section{Data for Alignment Guidance}
We leverage Grounded Segment Anything Model \cite{ren2024grounded} to obtain the masks of the target concepts to be focused on in the cross-attention layers.
For text-based segmetation (referring expression segmentation)  Grounded-SAM combines Florence-2 \cite{0004WXDHL00Y24}  and SAM \cite{kirillov2023segment} to obtain the masks for the given text.
Florence-2 takes a task instruction as input and generates results in the text form. 
Specifically for the referring expression segmentation, instruction ``Ground the object which is most related to the text input'' is provided. 
The segmentations are generated as polygons, with location tokens  $(x_0, y_0, \ldots, x_n, y_n)$ representing the vertices of the polygon in clockwise order.
The tokens and the image are provided to the SAM model to generate the target mask.
With this pipeline, during the pre-training stages (1--3) we generate the target masks for 77K images and their associated captions in the Cambrian 2.5M dataset.
Importantly, during fine-tuning, since the model takes image and a question prompt as input to generate the answer, we adhere to this framework and generate the segments based on the question for the given image. 
This instills in the model the ability to look at the right regions based on the question about the given image. 
For this phase, we utilize approximately 1\% (150K samples) of the Cambrian10M instruction fine-tuning dataset.

\section{Implementation Details}
For any stage, we use the learning rate of $1\mathrm{e}{-5}$ for all the modules including ConvNeXt, the projector, the cross-attention layers and the language model. 
We use Adam optimizer with the weight decay of 0.1, the warmup ratio of 0.03, $\beta_2$ is set to 0.95. 
Additionally, we train of each stage of a single epoch consistent with prior work on large vision-language models \cite{ge2024convllava}.

\section{Image Processing and Attention}
The input image to ConvNeXt is of size $768 \times 768$ yielding 576 tokens. 
We make an important observation the prior work \cite{ge2024convllava,liu2023visual} zero-pad the images to resize them to target resolution.
In our analysis we observe that for prior work without our attention guidance, the attention is focused on these padded regions. 
This might be an additional bottleneck for the vision-language models as they can easily ignore the vision features due to this inconsistency in the data.
The impact of image-preprocessing techniques in large models needs further investigation and is an important direction for future work.

\section{Evaluation Benchmarks}
We specifically evaluate on vision-centric benchmarks which take into account the visual information for visual question answering, suitable for detecting hallucination in multimodal setting.

\paragraph{HallusionBench \cite{guan2024hallusionbench}.}
The benchmark comprises 346 images paired with 1129 questions. 
The questions are framed in the yes/no format.
The questions also ask about objects which are not present in the image.
This allows for targeted evaluation for our goal of attention guidance to integrate visual information emphasizing that the model ``looks" at the image to perform the task.
\paragraph{OCRBench \cite{liu2024ocrbench}.} It evaluates the ability of VLMs to accurately detect and read text in the images.
In our model evaluation we focus on 475 images for the scene text centric VQA and key information extraction where images with text such as addresses, receipts, signs etc. are presented to the LMM and questions about the OCR content are asked.
\paragraph{CV-Bench \cite{tong2024cambrian}.}
This is a larger vision-centric benchmark containing 2638 manually-inspected examples.
This contains image-question pairs to evaluate 2D (spatial relationships, counting) and 3D (depth order, relative distances) understanding of the VLM. For this it uses, 
ADE20k, COCO and OMNI3D dataset benchmarks.

\paragraph{RealWorldQA \cite{realworldqa}.} 
This benchmark consists of 765 images, with a question and easily verifiable answer for each image.  This dataset also contains questions about spatial understanding in images. 
\paragraph{POPE \cite{li2023evaluating}.} Similar to HallusionBench, POPE is also inspired to evaluate VLMs for object hallucination. 
This also contains yes/no format of the questions about the absence or presence of objects in the image.

\section {Additional Qualitative Examples}
In Tab.~\ref{table:sup:Qualitative} and \ref{table:sup:attention_vis}, we present additional qualitative examples to show the performance of our model with attention guidance. 
Our model consistently performs better than Mobile-VLM V2 showing the advantages of our model in grounding its answers in the visual information. We support our results by visualzing the attention weights in Tab.~ \ref{table:sup:attention_vis} where our model is clearly able to look at the relevant regions for a given question. 
\begin{table*}[t]
\small
\begin{tabularx}{\linewidth}{@{}lXXX@{}}
\toprule
& {\includegraphics[width=0.7\linewidth,height = 1.5 cm]{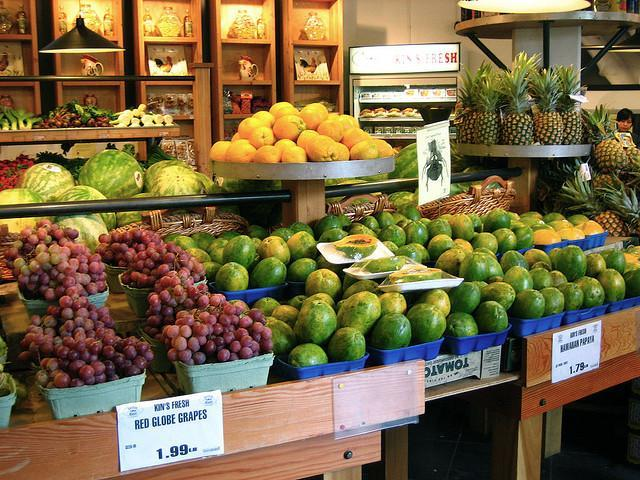}}
 & {\includegraphics[width=0.7\linewidth, height = 1.5 cm]{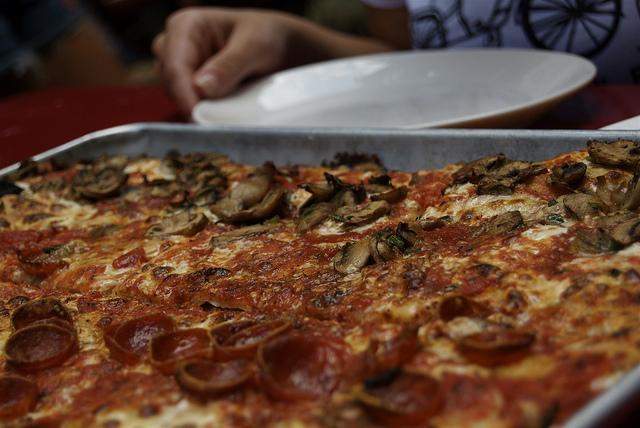}} &
 {\includegraphics[width=0.7\linewidth,height = 1.5 cm]{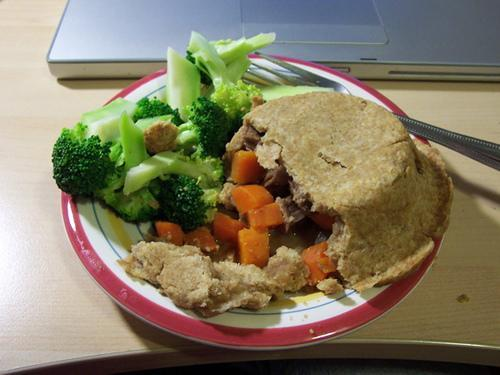}}
 \\
Prompt  & Is there a cup in the image? Please answer yes or no.  &  Is there a chair in the image? Please answer yes or no. & Is there a person in the image? Please answer yes or no.\\
Groundtruth & No & No & No\\
\midrule
MobileVLM-v2 & Yes, there is a cup in the image.&    Yes, there is a chair in the image.&  Yes, there is a person in the image. \\
\midrule
AGE-VLM & No, there is no cup in the image. The image primarily features fresh produce and does not include any cups or containers that could be used for drinking. The focus is on the display of fruits and vegetables, which are typically used for consumption rather than for drinking.
 &  No, there is no chair in the image. The image is focused on the pizza and the person's hands, with no chairs or furniture visible in the frame.
 & No, there is no person in the image. The image solely features food items on a plate, which are the main subjects. There are no people present in the frame.
\\
\midrule
\midrule

& {\includegraphics[width=0.7\linewidth,height = 1.5 cm]{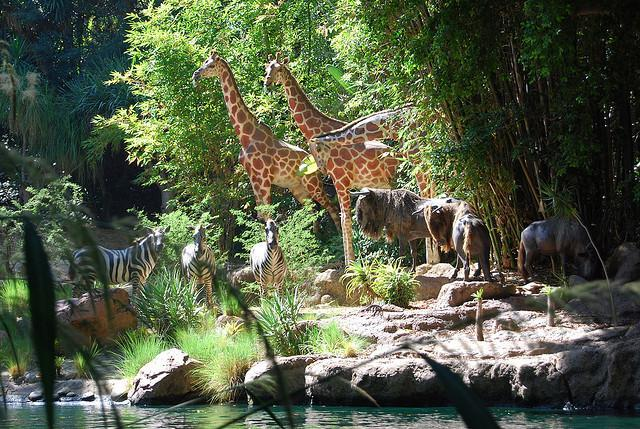}}
 & {\includegraphics[width=0.7\linewidth, height = 1.5 cm]{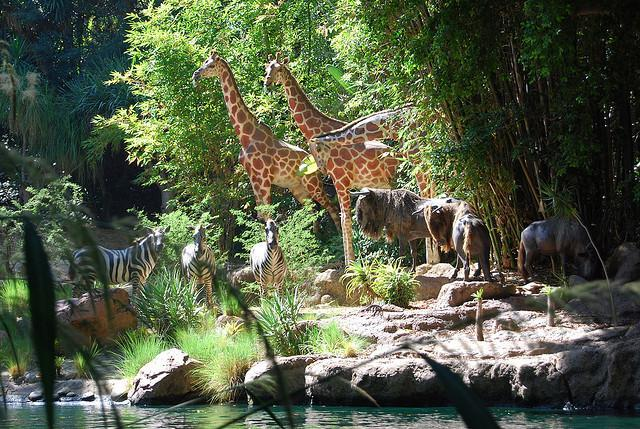}} &
 {\includegraphics[width=0.7\linewidth,height = 1.5 cm]{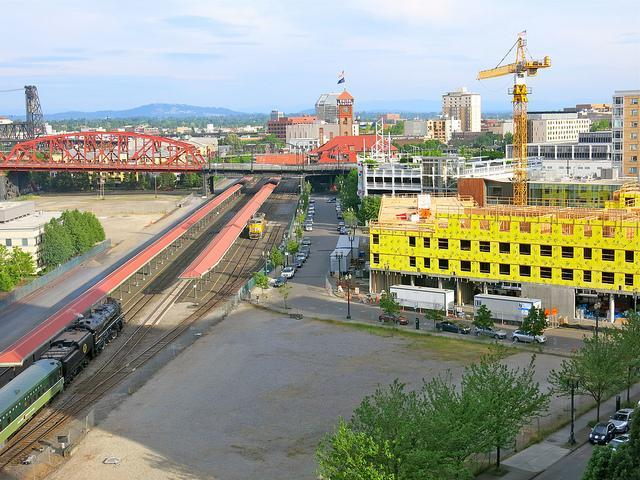}}
 \\
Prompt  & Is there a bicycle in the image? Please answer yes or no. & Is there a person in the image? Please answer yes or no. & Is there a person in the image? Please answer yes or no.\\
Groundtruth & no & no & no\\
\midrule
MobileVLM-v2 & Yes, there is a bicycle in the image. & Yes, there is a person in the image. & Yes, there is a person in the image. \\
\midrule
AGE-VLM & No, there is no bicycle in the image. The image is a natural scene with animals and vegetation, and it does not contain any man-made objects such as a bicycle.&   No, there is no person in the image. The image solely features wildlife, specifically giraffes and zebras, in a natural setting. There are no human figures or indications of human presence in the scene.&There is no person visible in the image.\\

\midrule
\midrule 
& {\includegraphics[width=0.7\linewidth,height = 1.5 cm]{2025-mahajan-neurips/sec/figures/supp/484.png}}
 & {\includegraphics[width=0.7\linewidth, height = 1.5 cm]{2025-mahajan-neurips/sec/figures/supp/1010.png}} &
{\includegraphics[width=0.7\linewidth, height = 1.5 cm]{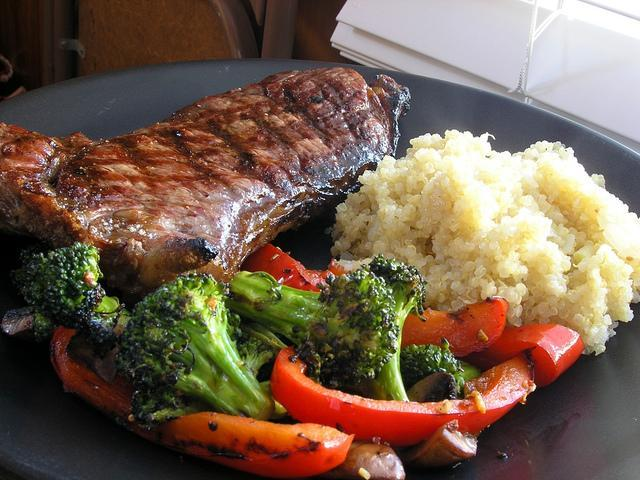}}
 \\
Prompt  &  Is there a backpack in the image? Please answer yes or no. & Is there a chair in the image? Please answer yes or no. &Is there a bowl in the image? Please answer yes or no.\\
Groundtruth & yes & no & no\\
\midrule
MobileVLM-v2 &No, there is no backpack in the image. The image features a black cat standing next to a bottle of wine. & Yes, there is a chair in the image. & Yes, there is a bowl in the image.  \\
\midrule
AGE-VLM & Yes, there is a backpack in the image. It is hanging on the wall to the right of the cat. &  No, there is no chair in the image. The image shows a table with a pizza and a drink, and there is no chair visible in the frame. & No, there is no bowl in the image. The food is presented on a plate, which is a shallow dish typically used for serving food rather than a bowl.\\
 \bottomrule
\end{tabularx}
\caption{Qualitative comparison of our approach against MobileVLM-v2 on vision-centric task.}
\label{table:sup:Qualitative}
\end{table*}

\begin{table*}[t]
\small
\begin{tabularx}{\linewidth}{@{}XXXX@{}}
\toprule
Image & Conv-LLaVA& CA-baseline & Ours \\ 
\midrule
 \raisebox{-51pt}{\includegraphics[width=\linewidth]{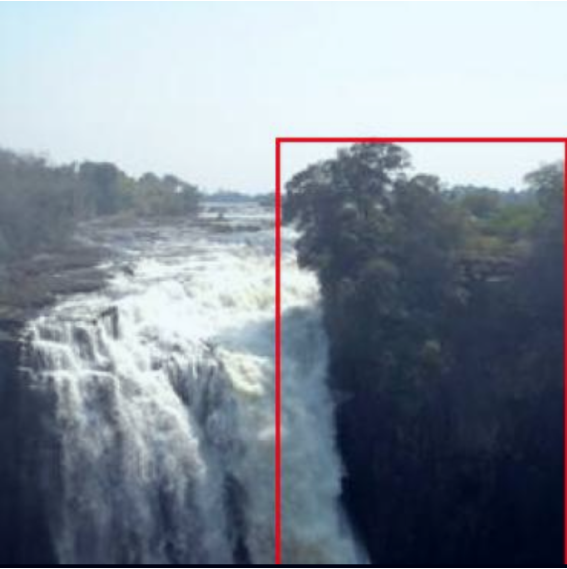}}
 & \raisebox{-51pt}{\includegraphics[width=\linewidth]{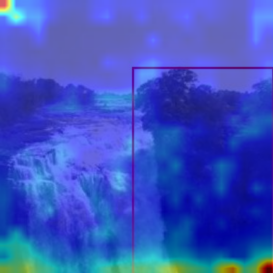}}
&\raisebox{-51pt}{\includegraphics[width=\linewidth]{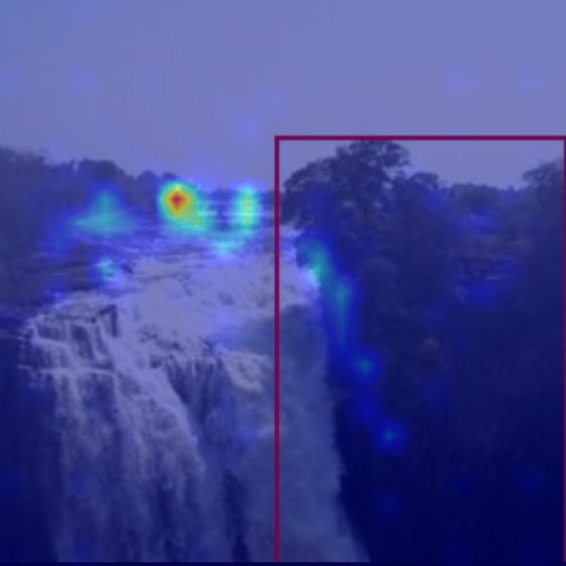}}

& \raisebox{-51pt}{\includegraphics[width=\linewidth]{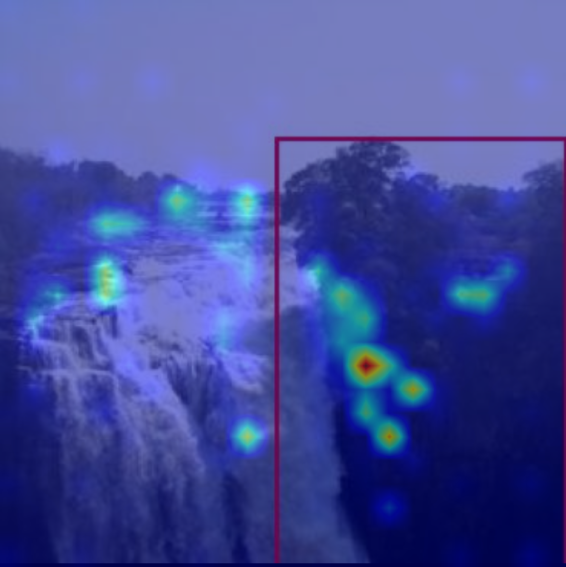}}\\
\multicolumn{4}{m{\textwidth}}{Considering the relative positions of the rocks (annotated by the red box) and the trees in the image provided, \newline where is the rocks (annotated by the red box) located with respect to the trees?}\\

\raisebox{-51pt}{\includegraphics[width=\linewidth]{2025-mahajan-neurips/sec/figures/Attention/433_gt.png}}
 & \raisebox{-51pt}{\includegraphics[width=\linewidth]{2025-mahajan-neurips/sec/figures/Attention/433_baseline_conv.png}}
&\raisebox{-51pt}{\includegraphics[width=\linewidth]{2025-mahajan-neurips/sec/figures/Attention/layer_0_data_433_baseline.png}}
& \raisebox{-51pt}{\includegraphics[width=\linewidth]{2025-mahajan-neurips/sec/figures/Attention/layer_0_data_433_ours.png}}\\
\multicolumn{4}{m{\textwidth}}{Considering the relative positions of the river water and the stone in the image provided, where is the river water located with respect to the stone?}\\
\raisebox{-51pt}{\includegraphics[width=\linewidth]{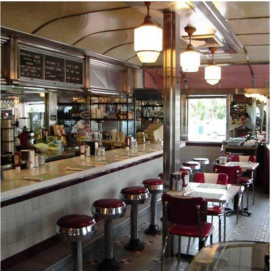}}
 & \raisebox{-51pt}{\includegraphics[width=\linewidth]{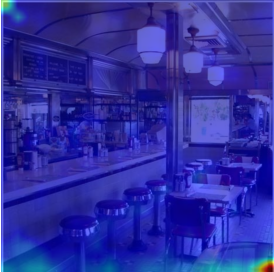}}
&\raisebox{-51pt}{\includegraphics[width=\linewidth]{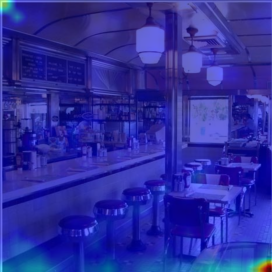}}
& \raisebox{-51pt}{\includegraphics[width=\linewidth]{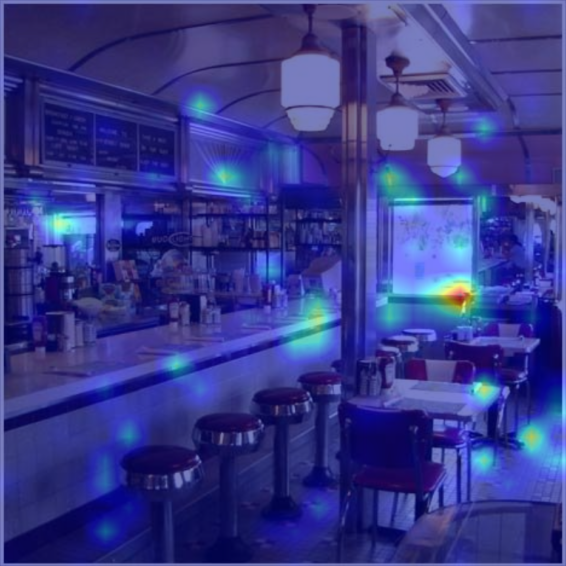}}\\
\multicolumn{4}{m{\textwidth}}{How many windows are in the image?}\\
\bottomrule
\end{tabularx}
\caption{\textbf{Attention visualization.} Our method looks at the right regions given the input image and the input text.}
\label{table:sup:attention_vis}
\vspace{-0.5 cm}
\end{table*}

\end{document}